\documentclass[preprint,12pt]{elsarticle}

\usepackage[ruled,vlined]{algorithm2e}%
\usepackage{float} %
\usepackage{graphicx}%
\usepackage{multirow}%
\usepackage{amsmath,amssymb,amsfonts}%
\usepackage{mathrsfs}%
\usepackage[title]{appendix}%
\usepackage{xcolor}%
\usepackage{textcomp}%
\usepackage{manyfoot}%
\usepackage{booktabs}%
\usepackage{array}%
\usepackage{listings}%
\usepackage{wrapfig}
\usepackage{lipsum} 
\usepackage{appendix}%
\counterwithin{figure}{section}%
\usepackage{tabularx}%
\usepackage{multirow}%
\usepackage{caption}%
\usepackage{parskip}%
\usepackage{url}
\usepackage{hyperref}

\journal{be decided}

\begin{document}

\begin{frontmatter}



\title{An Unsupervised Approach Towards Promptable Defect Segmentation In Laser Additive Manufacturing By Segment Anything}

\author[label1]{Israt Zarin Era}
\author[label1]{Imtiaz Ahmed\corref{cor1}}
\ead{imtiaz.ahmed@mail.wvu.edu }
\author[label1]{Zhichao Liu\corref{cor1}}
\ead{zhichao.liu@mail.wvu.edu}
\author[label1,label2]{Srinjoy Das\corref{cor1}}
\ead{srinjoy.das@mail.wvu.edu }
\affiliation[label1]{organization={Industrial and Management Systems Engineering, WVU},
            addressline={1306 Evansdale Dr.}, 
            city={Morgantown},
            postcode={26506}, 
            state={WV},
            country={USA}}
\affiliation[label2]{organization={School of Mathematical and Data Sciences, WVU},
            addressline={Armstrong Hall}, 
            city={Morgantown},
            postcode={26506}, 
            state={WV},
            country={USA}}

\cortext[cor1]{Corresponding author}
\begin{abstract}
Foundation models are currently driving a paradigm shift in computer vision tasks for various fields including biology, astronomy, and robotics among others, leveraging user-generated prompts to enhance their performance. In the Laser Additive Manufacturing (LAM) domain, accurate image-based defect segmentation is imperative to ensure product quality and facilitate real-time process control. However, such tasks are often characterized by multiple challenges including the absence of labels and the requirement for low latency inference among others. Porosity is a very common defect in LAM due to lack of fusion, entrapped gas, and keyholes, directly affecting mechanical properties like tensile strength, stiffness, and hardness, thereby compromising the quality of the final product. To address these issues, we construct a framework for image segmentation using a state-of-the-art Vision Transformer (ViT) based Foundation model (Segment Anything Model) with a novel multi-point prompt generation scheme using unsupervised clustering. Utilizing our framework we perform porosity segmentation in a case study of laser-based powder bed fusion (L-PBF) and obtain high accuracy without using any labeled data to guide the prompt tuning process. By capitalizing on lightweight foundation model inference combined with unsupervised prompt generation, we envision constructing a real-time anomaly detection pipeline that could revolutionize current laser additive manufacturing processes, thereby facilitating the shift towards Industry 4.0 and promoting defect-free production along with operational efficiency.
\end{abstract}




\end{frontmatter}


\section{Introduction}\label{sec1}
Real-time vision-based quality inspection \cite{gao2022review}  and control are essential for ensuring the reliability of the final product, particularly in industries such as aerospace and automotive where precision is non-negotiable for safety. Defect segmentation plays a crucial role in this process, as it aids in the root cause analysis of anomalies by precisely identifying and locating deviations from the expected standards. With the advent of Industry 4.0, \cite{oztemel2020literature}, defect segmentation has become integral to both preventive maintenance and corrective actions within the production pipeline to improve process efficiency and ensure high-quality products. However, it is generally challenging to carry out such tasks in industrial settings compared to medical or natural instances. Unlike medical or biological imaging domains where the data structure is relatively standardized \cite{azad2023foundational, mazurowski2023segment, wu2023medical} and labels are often available, industrial defects have diverse forms, low signal-to-noise ratios, and often insufficient labeling. This complexity necessitates the development of more adaptable and versatile segmentation techniques. Consequently, there is a critical need for approaches that are more customized and specifically tailored to industrial defect segmentation \cite{ji2023sam, moenck2023industrial}. In this paper, our discussion centers on a case study from \cite{kim2017investigation}, focusing on image segmentation for defects observed in cylindrical parts printed through the Laser-based powder bed fusion (L-PBF) process. Despite its wide adoption in various industries, L-PBF is susceptible to defects, such as porosity, cracking, and geometric distortions. Porosity is a very common defect in L-PBF due to lack of fusion, entrapped gas, and keyholes \cite{fu2022machine, kim2017investigation}. Porosity directly affects mechanical properties like tensile strength, stiffness, and hardness thereby compromising the quality of the final product. \cite{al2020effects}. Non-destructive tests like high-resolution X-ray Computed Tomography (XCT) are widely used to isolate and locate pores in the fabricated parts \cite{al2020effects, fu2022machine, kim2017investigation}. However, identifying and measuring inter-layer porosity from XCT data can be tenuous and challenging, and therefore to address this, machine learning-driven image segmentation has been adopted for such tasks which has significantly improved the overall process and product quality. A notable example is the use of U-Net for defect segmentation in Laser Additive Manufacturing (LAM). Originally designed for medical image segmentation, U-Net\cite{ronneberger2015u}, has emerged as a leading tool for segmentation in surface anomalies \cite{scime2020layer},  3D volumetric defects \cite{wong2021automatic}, and 3D geometry reconstruction \cite{ziabari2023enabling}, and has established itself as the benchmark in semantic segmentation of the LAM defects by achieving remarkable accuracy. However, in real-world scenarios, most of the LAM data are either unlabeled or limited in sample size. This directly conflicts with the conventional supervised training approach of U-Net and its variants.

In this work we address the aforementioned challenges by using the Segment Anything Model (SAM) \cite{kirillov2023segment} which is a Foundation model \cite{bommasani2021opportunities} developed by Meta for image segmentation tasks. SAM has been recognized for initiating a major shift in the field of image segmentation and can utilize user-specified hints also known as prompts \cite{jia2022visual, liu2023pre} to carry out specific segmentation tasks on new image datasets without prior training. However, a critical limitation of SAM is that it uses visual prompts, i.e., bounding boxes and points, which typically depend on human-generated annotations or interventions \cite{lin2020interactive}, which may not always be feasible to provide. To overcome this, in this paper, we introduce a simple yet novel framework for segmenting porosity in XCT images of specimens produced by L-PBF. To the best of the authors' knowledge, ours is the first work in the field of Laser Additive Manufacturing where a promptable foundation model is used for image segmentation in order to enable defect detection. Our primary contributions are listed as follows.  
\begin{itemize}
    \item We propose a novel porosity segmentation framework by incorporating the Segment Anything Model (SAM) using an unsupervised clustering-based approach for contextual prompt generation. 
    \item We utilize Bootstrapping to quantify the uncertainty of the model's performance across various prompt sets generated by our framework.   
    \item We provide detailed guidelines for obtaining accurate predicted masks from SAM's multi-prediction outputs in the context of XCT imaging of inter-layer porosity.  

\end{itemize}

The rest of the paper is organized as follows. In Section \ref{sec2}, we review the existing literature and highlight the research gaps. In Section \ref{sec3}, we discuss the dataset used in this work, and the foundation model Segment Anything. In section \ref{sec4}, we present our visual prompt generation framework, which employs an unsupervised clustering strategy to prompt the SAM model in order to perform inference. In Section \ref{sec5}, we demonstrate the predicted masks obtained from SAM using our framework, interpret the outcomes from the model, and measure the accuracy of our framework. In Section \ref{sec6}, we discuss the limitations and future potentials of the proposed framework. Next, in Section \ref{sec7}, we present our vision of constructing a real-time anomaly detection pipeline, channeling the full potential of such foundation models in manufacturing, especially in LAM incorporating IoT sensors. Finally, in Section \ref{sec8}, we summarize our findings and provide concluding remarks.

\section{Literature Review}\label{sec2}
We categorize the relevant literature into two main sections: prior studies on porosity segmentation in XCT images using statistical tools and machine learning models, and the advances of Vision Transformers which form the core of the SAM model in the field of image segmentation. These topics are further discussed in the following sections.  

\subsection{Porosity Segmentation on XCT Images}\label{subsec2.1}
Semantic segmentation \cite{yu2018methods} is a computer vision task that assigns a label to every pixel and helps to understand a pixel-wise mapping of the contents in an image. In the rapidly advancing field of Artificial Intelligence (AI), the domain of image segmentation within manufacturing continues to undergo significant transformations. 
For porosity segmentation on the metallography images of parts fabricated via LAM, \cite{garcia2020image} performed a Hessian matrix-based image analysis to select the candidate pores. In \cite{malarvel2021autonomous} autonomous techniques were proposed for weld defect detection using machine learning classifiers such as Support Vector Machines (SVMs) and Feed Forward Neural Networks. In \cite{singh2021computer} Principal Component Analysis (PCA) in conjunction with K-means clustering was used to segregate features of pores and fractures in XCT images of sandstones. \cite{chauhan2014comparison} used K-means clustering, Fuzzy C-means, and Self Organizing Maps to perform the porosity segmentation in XCT images of the petrophysical rocks to classify pores using a Neural Network and a Support Vector Machine Classifier. However, traditional machine learning techniques like Support Vector Machines and Random Forests, often combined with image thresholding methods such as Bernsen \cite{bernsen1986dynamic} and Otsu Thresholding \cite{otsu1975threshold}, have been superseded recently by Deep Learning based algorithms. In \cite{alqahtani2020machine} Convolutional Neural Networks (CNNs) were trained to estimate various physical properties of porous media by utilizing micro-computed tomography (micro-CT) X-ray images as input data.  Otsu thresholding was combined with a Convolutional Neural Network in \cite{gobert2020porosity} to automatically segment porosity in XCT images of metallic AM specimens. Similarly, \cite{bellens2022evaluating} used the segmentation results of Otsu Thresholding as ground truth labels for training a U-net \cite{ronneberger2015u}. \cite{wong2021automatic} proposed a volumetric porosity segmentation technique that employs U-Net based on a ResNet backbone with Bernsen Thresholding being used to generate the ground truth. While such studies have utilized diverse statistical thresholding techniques to generate ground truth labels for XCT images for supervised model training, setting these thresholds across images from different samples can be challenging. In addition, most of the aforementioned works utilize traditional supervised learning methods to train the models for performing defect segmentation. While this can yield high prediction accuracy, such methodologies also require large labeled datasets and incur significant computational overhead for training. These challenges can be addressed by Transfer learning \cite{van2014transfer} which is a machine learning technique that can leverage the knowledge learned from training on one task to perform different but related tasks. Such pre-trained models can be incrementally trained (fine-tuned) on new unseen target data to obtain high performance. Nevertheless, Transfer learning, even when utilizing pre-trained models such as U-Net, still necessitates partial training of the decoder using labeled data \cite{ouidadi2023defect}, which might not be a feasible option for many real-life settings. In addition, pre-trained models are unlikely to align smoothly with hitherto unseen target data owing to factors such as domain shift and this can greatly increase the complexity of fine-tuning.

\subsection{Evolution of Vision Transformers in Image Segmentation}\label{subsec2.2}
Starting with their remarkable success in natural language processing (NLP) \cite{brown2020language}, foundation models are now being widely adopted for various tasks in computer vision \cite{haas2023learning, ramesh2021zero}. These are machine learning models that are trained using self-supervised learning techniques on vast amounts of unlabeled data and can perform inference on unseen target data with little or no fine-tuning. Foundation models have proven to be highly versatile and in general, they can be applied to a broad range of tasks without requiring specific training for individual applications \cite{bommasani2021opportunities}. Most foundation models in computer vision are built using the Vision Transformer Encoder and Decoder architecture (ViTs) \cite{ma2023towards, vaswani2017attention}. Compared to traditional CNNs \cite{alzubaidi2021review} which are Deep Learning models that have been extensively used for various tasks in computer vision,  ViTs exhibit certain distinct advantages. CNNs are adept at inference tasks such as image recognition and segmentation by generating feature maps that emphasize local image details and exhibit shift invariance \cite{singla2021shift}. However, their performance often suffers when there are significant changes in the orientation of objects or patterns in the image. ViTs, on the other hand, rely on an attention mechanism, enabling them to capture global context and long-range dependencies in addition to local information \cite{dosovitskiy2020image}. Multi-headed Attention in ViTs further enhances their capability by allowing different heads to focus on various aspects of the input, thereby producing a comprehensive representation combining both local and global contexts \cite{vaswani2017attention}. These models outperform CNNs in tasks like image classification, object detection, and semantic segmentation by simultaneously considering information from various parts of the input image \cite{mauricio2023comparing}. Several Vision Transformer architectures have been proposed recently for computer vision tasks like image classification, dense prediction, and semantic segmentation including Swin Transformer \cite{liu2021swin}, Segmentation Transformer (SETR) \cite{zheng2021rethinking} and SegFormer \cite{xie2021segformer}. Efficient Transformer \cite{xu2021efficient} is another variant that has been developed specifically for performing semantic segmentation on high-resolution remote sensing images. ViT architectures like COTR (consisting of convolutional and transformer layers) \cite{shen2021cotr}, SegTran, (Squeeze-and-Expansion Transformer) \cite{li2021medical}, SWIN-Unet \cite{cao2022swin} have also been applied in the medical domain for 2D and 3D medical image segmentation. These hybrid Convolutional-Transformer architectures attempt to combine the strengths of CNNs with Transformer models.

As task-agnostic foundation models increase in popularity, SAM has emerged as a solution for generalized image segmentation. The architecture of SAM has three key components: an image encoder as the initial processing step, a prompt encoder that accepts both sparse prompts (bounding boxes, points, texts) and dense prompts (masks), and a lightweight mask decoder that generates segmentation masks using image embedding, prompt embedding, and an output token. For image segmentation tasks SAM has been shown to produce excellent performance in several fields such as medical imaging \cite{shi2023generalist, zhang2023customized, ma2024segment}, camouflaged object detection \cite{tang2023can},  moving object tracking \cite{yang2023track}, pseudo label generation \cite{ahmadi2023application, chen2023learning} and image captioning \cite{wang2023caption} among others. In some of these applications, SAM has been found to face challenges in real-world tasks without fine-tuning \cite{ma2024segment}. In the field of additive manufacturing, SAM has also been reported to face difficulties, especially for cases of unusual instances \cite{ahmadi2023application, ji2023sam}. In our problem of porosity segmentation, we find that directly applying SAM to the XCT images of the defects does not produce optimal results.  In this work, we aim to address such limitations which can unlock the full potential of SAM for image segmentation in additive manufacturing.
 
Our research reveals that integrating contextual prompt generation into the SAM framework provides weak supervision \cite{zhou2018brief} for segmentation, thereby significantly enhancing the model's performance and a marked improvement in the quality of the final result. Our framework also avoids the overhead of supervised training or direct fine-turning both of which requires a large number of labels. As demonstrated by our results which are discussed later, our unsupervised approach for prompt generation not only achieves high-accuracy predictions but also simplifies the challenges associated with manual prompt annotation. This proves the effectiveness of our framework as a more generalized approach for performing industrial defect segmentation in real-world settings.

\section{L-PBF Dataset and SAM Model}\label{sec3}
In this section, we briefly describe the dataset and discuss the key components of the Foundation model used in this study i.e., the Segment Anything Model. 
\subsection{Dataset}\label{subsec3.1}
\begin{figure}[h!]%
\centering
\includegraphics[width=1\textwidth]{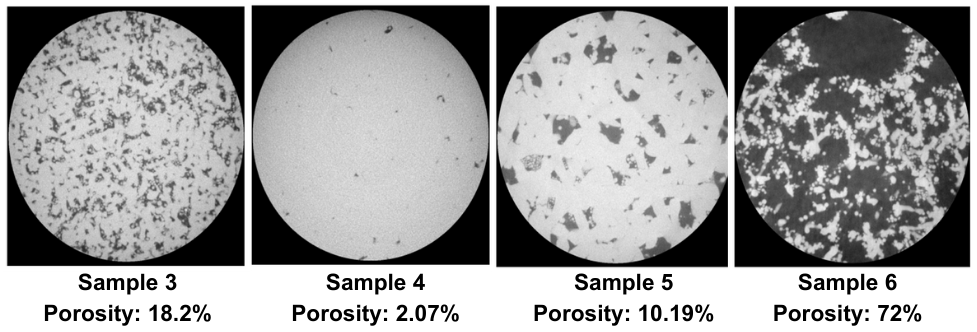}

\caption{XCT images showing the surface topology of defects of two CoCr alloy cylindrical discs; from left to right respectively from \cite{kim2017investigation}. }\label{fig3.1}
\end{figure}

The data presented in this study is derived from \cite{kim2017investigation}, where Cobalt-chrome alloy specimens, containing elements such as Co, Cr, Mo, Si, Mn, Fe, C, and Ni, were fabricated through L-PBF processes. The study involved high-resolution X-ray computed tomography (XCT) scans on the L-PBF manufactured CoCr specimens to explore the effects of varying processing parameters, specifically scan speed and hatch spacing on characteristics such as pore structure formation and overall porosity. The dataset consists of 8-bit grayscale images of 2D slices of the top surface of the cylindrical 3D samples. For our investigation, we use the data from four specimens i.e., Samples 3, 4, 5, and 6 respectively (using the same terminology as the original paper where this dataset is described \cite{kim2017investigation}). We select these four samples based on their variations in volume, size, and distribution of porosity as presented in Figure \ref{fig3.1} and Table \ref{tab1} respectively. 

Example images of Samples 3 through 6, as depicted in Figure \ref{fig3.1}, reveal significant differences in pore structures attributed to varying processing parameters. Note that pores are considered as a function of both hatch space and scanning speed \cite{kim2017investigation}. Based on the data in Table \ref{tab1}, it can be seen that Sample 4 exhibits less porosity with smaller closed pores compared to other samples, without trapped powder particles on account of the lowest hatch space. Samples 3 and 5 exhibit distinct pore structures despite having a similar amount of 
 (\%) porosity. Sample 3 demonstrates comparatively smaller and more connected pores compared to Sample 5 with larger and visually disconnected pores. Both of these two samples have unmelted particles trapped inside the pores due to higher hatch space. Sample 6, with a porosity of 72\%, displays a disconnected solid print part in 2D, primarily due to its high porosity.

\begin{table}[h]
\captionsetup{width=1\textwidth}
\centering
\small
\begin{tabularx}{\textwidth}{@{}c*{4}{>{\centering\arraybackslash}X}@{}}
\hline
\textbf{Samples} & \textbf{Image size (px)} & \textbf{Scanning Speed (mm/s)} & \textbf{Hatch space (mm)} & \textbf{Porosity (\%)} \\
\hline
\textbf{Sample 3} & (988, 1013) & 3200 & 0.1 & 18.2  \\
\textbf{Sample 4} & (984, 1010) & 800 & 0.2 & 2.07 \\
\textbf{Sample 5} & (984, 1010) & 800 & 0.4 & 10.19 \\
\textbf{Sample 6} & (984, 1013) & 3200 & 0.4 & 72.0 \\
\hline
\end{tabularx}
\caption{Porosity information of the XCT images of the samples at varying process parameters.}
\label{tab1}
\end{table}
\newpage
\subsection{Segment Anything Inference Model}\label{3.2}

\begin{figure}[h]%
\centering
\includegraphics[width=1\textwidth]{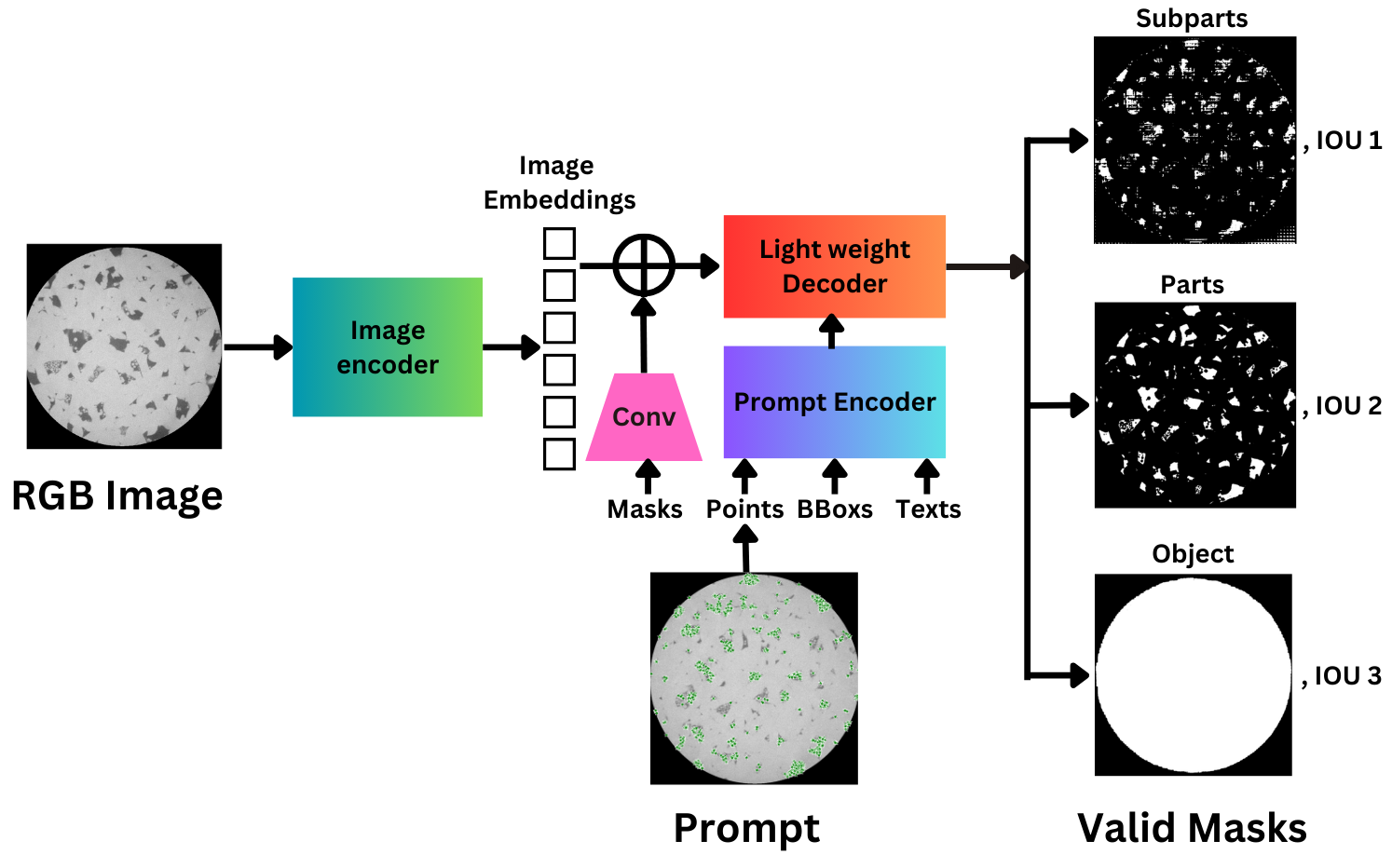}
\caption{The overview of SAM's architecture redrawn from the original paper \cite{kirillov2023segment}.}\label{fig3.4}
\end{figure}

The architecture of SAM consists of an image encoder, a prompt encoder, and a mask decoder (Fig. \ref{fig3.4}). The main components are discussed as follows.  

\textbf{ Image Encoder: } 
The Image Encoder expects RGB images and the input image is rescaled to size 1024 \(\times\) 1024  along with padding.  The image encoder employs a Masked Autoencoder (MAE); a pre-trained Vision Transformer (ViT) \cite{he2021masked}, which outputs a downscaled embedding of the input image, reduced by a factor of 16 in both dimensions. Initially, the image is divided into non-overlapping patches using a 16×16 kernel with a 16 by 16 stride, reducing the dimensions from 3 channels with a resolution of 1024×1024 to 768 channels with a resolution of 64×64. Subsequently, these patches are processed through a stack of 12 Transformer blocks, each comprising attention heads and multilayer perceptron (MLP) heads. In the encoder architecture's neck, a 1×1 convolution is applied to reduce the channel size from 768 to 256. This is followed by a 3×3 convolution, which produces the final output of the image encoder with dimensions of 256 channels and a resolution of 64×64. Following the notations from \cite{jia2022visual}, an RGB image \(I\) is converted into image embeddings, \(e^j\) by a convolutional operation, CONV as below.  

\begin{equation}
    e^{j} = \text{CONV}(I), \text{ where } I \in \mathbb{R}^{3\times 1024\times 1024}, e = \{e^j \in \mathbb{R}^{64\times 64}, 1 \leq j \leq 768\}.
\end{equation}\label{eq1}

Then the final output embedding from the image encoder \(Encoder^{I}\) is; 
\begin{equation}
    E = \text{Encoder}^I(e), \text{ where } E = \{e^q \in \mathbb{R}^{64 \times 64}, 1 \leq q \leq 256\}.
\end{equation}\label{eq2}
\textbf{Prompt Encoder:}
The prompt encoder accommodates different types of prompts: sparse prompts (bounding boxes, points) represented through positional embeddings combined with image embeddings, and dense prompts (masks) embedded via convolution and element-wise summation using the image embeddings and text prompts \cite{kirillov2023segment}. In our work, we use sparse prompts in the form of points to perform image segmentation. For point prompts, the representation of each point comprises a positional encoding of its location which is combined with embeddings denoting the location class of the point. 
These are finally transformed into 256-dimensional vectors to align with the image embeddings generated by the image encoder as described previously. From the prompts \(p\) and labels \(l\) provided by the user, the prompt embeddings generated by the prompt encoder, \(Encoder^{P}\), are: 
\begin{equation}
    [x^*, P] = \text{Encoder}^P([l, p]),  
     \text{where, } P \in \mathbb{R}^{256}
\end{equation}\label{eq3}
Here, \(x^*\) represents a learned embedding indicating the location class of a point which can be either the foreground or background, and is inserted along with the prompt embeddings \(P\), which are then called `tokens' in the original SAM paper \cite{kirillov2023segment}. 

\textbf{Mask Decoder}: 
During the inference phase, the mask decoder takes both the 256-dimensional image embeddings and the prompt tokens as input, producing masks as output. The mask decoder has two transformer layers and in each layer, a sequence of operations is executed as follows. In the first layer, self-attention is initially applied to the tokens. Subsequently, cross-attention is applied to the image embeddings using the tokens as queries. Next, a point-wise MLP updates each token. Finally, cross-attention is applied to the tokens using the image embeddings as queries. All the multi-head attention layers consist of 8 single attention heads. The following layers receive these updated image embeddings and tokens to improve the representation. Throughout these operations, residual connections, layer normalization, and dropout regularization are employed for stability and robustness. Moreover, the original prompt tokens along with their positional encodings, are reintroduced into the updated tokens during attention operations, enabling the model to have a comprehensive understanding of the spatial characteristics and semantic attributes of the original prompts. 

Following the decoder layers, the enhanced image embedding undergoes an upscaling process facilitated by two transposed convolutional layers, followed by another token to the image query attention block. Then the updated output tokens pass through three MLP heads computing the mask foreground probability at each image location. A spatial dot-product is executed between the upscaled image embedding and the outcome of the multiple MLP heads to predict the final masks. On account of this, it is possible to simultaneously obtain multiple masks for different regions of interest denoting sub-parts, parts, or the whole object in a single image from the same set of prompts. Moreover, the decoder also generates a confidence score, estimating the IoU for each mask to assess the quality of masks. 
For the \(N\) number of \(Layer\)s in a transformer block, 
\begin{equation}
    [x_i^*, E_i] = \text{Layer}_i([x_{i-1}^*, P_{i-1}, E_{i-1}]), \quad \text{where } 1 \leq i \leq N, \quad N = 2
\end{equation}\label{eq4}
\noindent Here, \(x^*\) represents the learnable class [cls] token \cite{dosovitskiy2020image}, and \(P\) and \(E\) as the prompts and image embeddings respectively.

The final output of the Mask Decoder of SAM, \(Decoder^{M}\) is denoted as below: 
\begin{equation}
    (M, S) = \text{\(Decoder^{M}\)}([x^*, P, E])
\end{equation}\label{eq6}
\noindent Here, \(M\) denotes the final predicted mask and \(S\) is the predicted IoU score by the model. \(x^*\) denotes the learnable class token, \(P\) denotes the prompts from prompt encoder \(Encoder^P\) and \(E\) denotes the image embeddings from image encoder \(Encoder^I\).

Since a single input prompt can correspond to multiple valid masks, SAM provides three valid final masks simultaneously instead of one to avoid ambiguity i.e., Subpart, Part, and Whole object. Masks returning the subparts of the object typically cover smaller regions or fractions of regions as compared to regions covered by masks returning parts of the whole object. SAM is prompted with a $32\times32$ regular grid of points for each input image, yielding a set of masks for each point. 
In addition to the masks, SAM also returns an estimated IoU score \cite{kirillov2023segment} based on its prediction for each of the masks. It should be noted that during inference, the estimated IoU scores obtained with each predicted mask are not calculated using our ground truth, but rather are predicted by SAM during inference to rank them for specific applications.

For porosity segmentation, we observe that masks identified as parts are the most suitable for our application and achieve the highest DSC score (refer to Section \ref{subsec5.1}) for how DSC scores are calculated) among the three masks. Sub-part masks typically contain comparatively less information about the pores than part masks, and the masks segmenting the entire circular surface of the samples as the whole object are not useful in our case. However, it is observed that in certain samples, both the part and whole object masks return identical outcomes. In such cases, the sub-part masks are deemed to be more precise and also verified in our experiments with the generated ground truth labels. Since we opt for an end-to-end unsupervised protocol in our framework, we select the most appropriate masks based on their ranks by the estimated IoU scores from SAM. 

The three masks representing various parts of the object are derived using different thresholding limits, although the specific details are not explicitly provided in the original paper. It's important to note that the masks are consistently indexed in the order of their ranks determined by SAM's predicted IoU. Typically, sub-part masks exhibit the lowest IoU scores, while whole object masks tend to have the highest IoU scores among the three. We find that for all the samples of our dataset, the predicted IoU for any mask returning the whole object is usually higher than 0.88-0.90. Thus we use it as a threshold to select the corresponding mask as shown in Algorithm \ref{alg3} in Section \ref{subsec4.2}. The strategy is to always select the mask that identifies the parts, unless its IoU score surpasses the threshold, indicating that it also returns the entire object. In such cases, the sub-part mask is selected as the final predicted result. 

\section{Proposed Framework}\label{sec4}
\begin{figure}[h]
    \centering
    \includegraphics[width=1\textwidth]{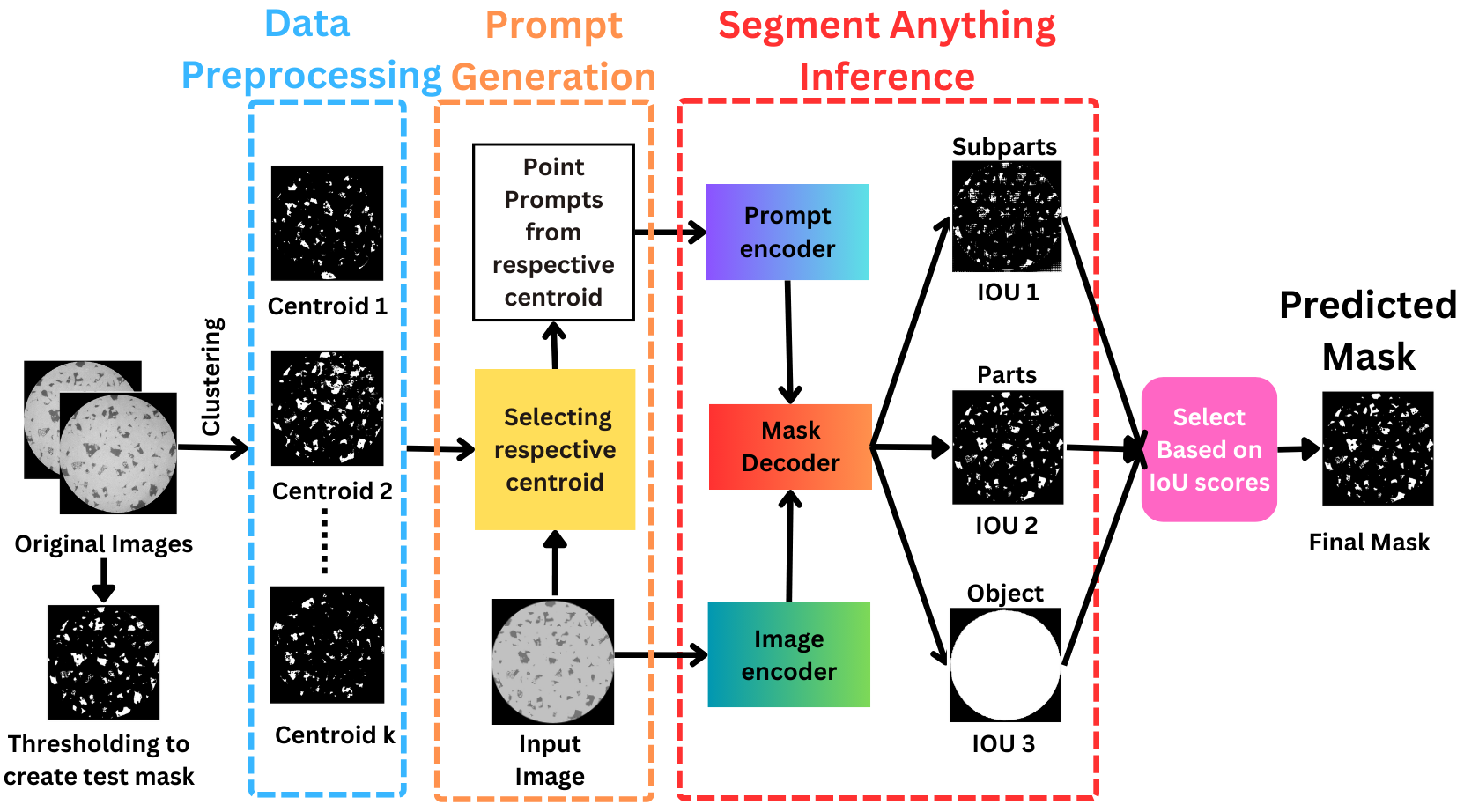}
    \caption{Our proposed framework of unsupervised prompt generation with Segment Anything Model.}
    \label{fig3.2}
\end{figure}

Our proposed framework in Figure \ref{fig3.2} consists of three primary stages: data preprocessing, prompt generation, and performing inference using SAM. Initially, in the data preprocessing phase, we perform unsupervised clustering to group similar images and obtain the centers of the respective clusters i.e., the centroids. Using the original images we also create reference binary masks using thresholding techniques to evaluate the performance of the framework after segmentation has been performed by SAM. Subsequently, we utilize these centroids as sources to generate prompts for individual images based on their respective assigned clusters. Finally, in the last step, we denoise the images and then apply them along with their corresponding prompts into the SAM Inference Model to obtain the segmented porosity masks. These steps are discussed in detail in the following sections. 
 \subsection{Data Pre-processing}\label{subsec4.1}
 In this section, we provide a detailed discussion of the data preprocessing steps. These steps involve the unsupervised clustering of the data to generate centroids, the preprocessing of the original RGB images, the prompt generation using the centroids, and finally, the creation of the Reference Binary masks.

First, we select a set of images from each of the samples discussed in Section \ref{subsec3.1} and perform unsupervised clustering. The goal of this is to determine the centroid images representing each individual cluster of XCT images, which will be utilized later to generate prompts for SAM.  In this work we use two Partitional Clustering \cite{madhulatha2011comparison} techniques namely K-means clustering \cite{ding2004k} and K-medoids \cite{park2009simple} clustering to determine the centroid images of the clusters. This is owing to their distinct ways of representing the clusters and the use of different distance metrics to quantify similarities among the data points. K-means is a widely used unsupervised clustering method due to its simplicity yet effectiveness in diverse data instances, but it is sensitive to outliers. On the other hand, K-medoids clustering, which employs an actual data point as the centroid of a given cluster, is less susceptible to noise or outliers. However, it may overlook subtle yet significant cluster information from the rest of the data. Therefore, we employ both unsupervised clustering techniques in our experiments for the unsupervised prompt generation strategy.  

\textbf{K-means Clustering}: 
The K-means algorithm begins with the random initialization of K cluster centroids. It then iteratively assigns each data point to the nearest centroid and updates the centroids by computing the mean of the data points assigned to each cluster. This iterative process continues until convergence, often determined by a minimal change in cluster assignments or centroids. The primary objective of the algorithm is to minimize the total squared distance between each data point and its assigned centroid. K-means clustering is susceptible to the initial positions of centroids and may converge to a local minimum. Multiple iterations are usually performed to address this issue and determine the optimal assignment of the original data points to individual clusters. The equation for K-means clustering is as shown below:
\begin{equation}\
J_K = \underset{U}{\text{argmin}} \sum_{k=1}^{K} \sum_{i \in U_k} (x_i - \bar{\mu}_k)^2
\end{equation}\label{eq7}
Here, \(J_K\) is the objective function, \( x_i \) is the data matrix \( (x_1, \ldots, x_n) \), the centroid of cluster \( U_k \) is calculated as \( \bar{\mu}_k = \frac{1}{|U_k|} \sum_{i \in U_k} x_i \), \( U \) is the collection of K number of clusters, and \( |U_k| \) is the number of points in cluster \( U_k \).

\textbf{K-medoids Clustering}: 
The K-medoids algorithm, an adaptation of K-means clustering, aims to divide a dataset into K clusters, with each cluster represented by one of its data points known as the medoid. 
K-means clustering estimates the centroids of each cluster based on the mean of the data points assigned to the cluster. In contrast in K-medoids, each cluster is represented by one of the data points in the cluster which is the medoid.
The algorithm initializes by selecting K medoids, often randomly or via a heuristic, and the rest of the data are assigned to the cluster whose medoid is closest to them. Subsequently, it iteratively updates the medoids by swapping them with other data points and recalculating the total cost based on the selected distance function. The process continues until convergence when the data is divided into K number of clusters with the respective medoids. The equation for K-medoids clustering is as shown below:
\begin{equation}\label{eq8}
J_K = \underset{U}{\text{argmin}}\sum_{k=1}^{K} \sum_{i \in U_k} D(x_i, m_k) 
\end{equation}
Here, \( J_K \) represents the objective function, \( x_i\) is the data matrix \( (x_1, \ldots, x_n) \), and \( D(x_i,m_k) \) is the distance between data point \( i \) and the medoid \( m_k \) of cluster \( U_k \). The medoid \( m_k \) is a real data point existing in the dataset \( X \). K-medoids algorithm can successfully leverage various distance functions \( D(x_i,m_k) \), including the Euclidean distance \cite{danielsson1980euclidean}, as well as alternatives like the Manhattan Distance \cite{mohibullah2015comparison}, Mahalanobis Distance \cite{mclachlan1999mahalanobis} Dynamic Time Warping Distance \cite{muller2007dynamic} and kernel-based distances \cite{das2021kernel}. In our investigation, we utilize both Euclidean distance and Dynamic Time Warping distance in the K-medoids clustering of the XCT images. These distance functions are briefly described below.

\textbf{Euclidean Distance (ED)} is the most common distance metric used in statistics, machine learning, and image processing. Euclidean distance measures the straight line distance between two points in a multidimensional space to capture their closeness. While ED is very simple to calculate, one of its principal disadvantages is its reduced effectiveness in high-dimensional spaces due to the curse of dimensionality. In such spaces, the distance between any two points tends to become similar, diminishing the ability to distinguish between them. Additionally, ED is sensitive to the scale of the data, computationally expensive for large high-dimensional datasets \cite{faber2001pros} and can be affected by redundant or irrelevant features. The equation for ED is as shown below:  

\begin{equation}
    D(A,B) = \sqrt{\sum_{i = 1}^{n} (A_i - B_i)^2} 
\end{equation}\label{eq9}

\noindent Here, \(D\) is the ED distance between two points \(A\) and \(B\) in n-dimensional space where \(A_i\) and \(B_i\) are points of Cartesian coordinates and \((1\leq i \leq n)\)  and \(n \in Z^+\).
  
\textbf{Dynamic Time Warping Distance (DTW)} is a distance measure used to estimate the similarity between two sequences. Unlike the Euclidean distance, which assumes a one-to-one correspondence between elements of the sequences, DTW allows for the two sequences to have different numbers of points. The equation for DTW is as shown below:   
\begin{equation}
    D(a,b) = \underset{\Pi}{\text{min}}\sqrt{\sum_{i \in \Pi}^{m}\sum_{j \in \Pi}^{n} (a_i - b_j)^2}
\end{equation}\label{eq10}
\noindent Here, \(D\) is the DTW distance between two sequences \(a\) and \(b\) with \(m\) and \(n\) lengths respectively. \(a_i\) and \(b_j\) are the \(i\)th and \(j\)th elements in the sequences \(a\) and \(b\) respectively. A (m,n)-warping path \(\Pi = (\Pi_1, \ldots, \Pi_L)\) establishes an alignment between two sequences \(a = (a_1, a_2, \ldots, a_m)\) and \(b = (b_1, b_2, \ldots, b_n)\) by assigning the element \(a_{i}\) of \(a\) to the element \(b_{j}\) of \(b\). Here \(L\) is the length of the warping path \(\Pi\). The boundary condition of DTW distance ensures that the first and last elements of the two sequences \(a\) and \(b\) are aligned with each other \cite{muller2007dynamic}.

\textbf{Rationale for using Partitional Clustering}: 
Our framework is designed to consider the layer-by-layer printing approach of the LAM process and we treat the data collected as the function of the layers in order. Moreover, the overall porosity of every sample is a function of its height in the build direction \cite{kim2017investigation} and the XCT images are collected in the same direction. Given this specific characteristic of the manufacturing sequence, we opt to use partitional clustering techniques which can provide a mean or center point accurately representing the central tendency of the clusters at each layer.

\begin{figure}[h]%
\centering
\includegraphics[width=1\textwidth]{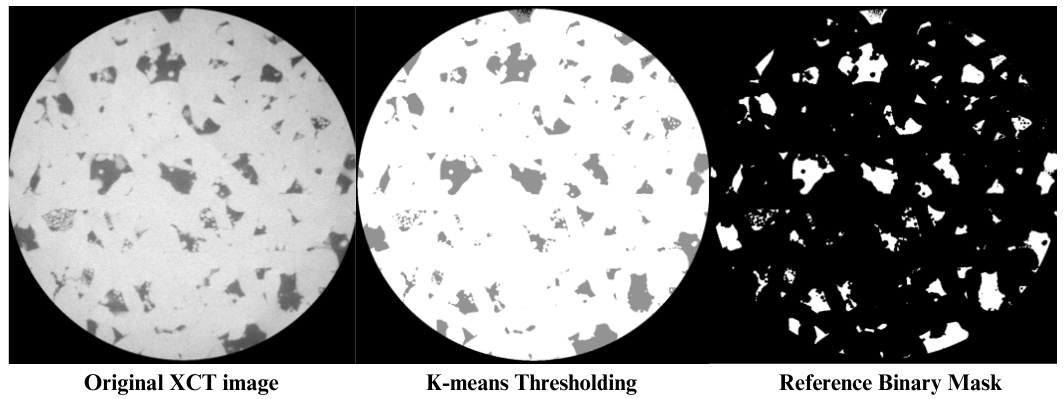}

\caption{K-means thresholding and Binarization of the XCT image to create Reference Binary masks; Original XCT image, K-means Thresholding, and Final Binary Mask from left to right respectively.}\label{fig3.3}
\end{figure}

\textbf{Reference Binary mask generation by Thresholding}: 
To evaluate the performance of our segmentation framework, it is required to have labels that distinguish defects from the remaining components in the images. The dataset utilized in this study does not contain pre-existing labels, and generating them manually for such a large data set is not a viable option. In our approach, we do not rely on these ground truth labels for training or fine-tuning SAM. However, we require them as a reference to assess the quality of the segmentation masks generated by SAM. 
We do this by applying K-means clustering \cite{ouidadi2023defect, panwar2016image} on the pixels of the XCT images with median filtering \cite{1163188} followed by binary thresholding to create the Reference Binary masks as demonstrated in Figure \ref{fig3.3}. We then compare these generated labels with the model's outputs and evaluate the performance of our framework. This is described as follows.

\begin{algorithm}[H]
\SetAlgoLined
\textbf{Input:} Image $I$; \\
\textbf{Output}: Reference Binary mask image $R$; \\

\textbf{Steps}: \\
Perform K-means clustering on the pixels of image $I$ with $K = 3$; \\
Run median filtering for denoising; \\
Obtain the centroid values $c_1$, $c_2$, and $c_3$; \\
Select a threshold ($T = c_2 $) for pixel binarization; \\ 
\For{each pixel $p$ in image $I$}{
    \If{$p > T$}{
        Set pixel value to 0\;
    }
}
\For{each pixel $p$ in image $I$}{
    \If{$p > 0$}{
        Set pixel value to 255\;
    }
}
\caption{Binary Thresholding}\label{alg1}
\end{algorithm}

Each XCT image across all samples exhibits three primary components: a dark background, a bright solid surface, and dark porous regions. Grayscale images typically range from 0 to 255, where 0 represents a completely black pixel and 255 represents a completely white pixel. In the K-means thresholding process, we set K = 3 to segregate pixels into these three categories within the XCT images. After obtaining centroid values, we observe that pixels corresponding to the dark background have low values close to 0, while those representing the solid surface have higher values closer to 255. The pixels related to the porous zones fall within the mid-range of this spectrum. Selecting the middle centroid value as the threshold, we proceed with binary thresholding as outlined in Algorithm 1 to generate binary masks where pores are depicted as the foreground (Class `1') and the rest of the image forms the background (Class `0').
This Binary Thresholding method performs well for Sample 3, Sample 5, and Sample 6. However, in the case of Sample 4, it is sensitive to the noise and complex variations in the pixel values observed in this sample. Therefore, in this case, we have made some manual adjustments to label the locations of pores in the images corresponding to this specific Sample.

\subsection{Prompt Generation}\label{subsec4.2}
Prompts can be defined as creating a task-specific information template. In natural language processing (NLP), prompting transforms the downstream dataset into a language modeling problem \cite{bahng2022exploring}. This approach allows a pre-trained language model to adapt to a new task without modifying its parameters. For computer vision tasks, visual prompts were first conceptualized in \cite{bahng2022exploring} using analogous ideas from NLP. Point prompts, bounding boxes, and masks fall into the broader categories of visual prompts and provide structured annotations in the data to guide the model in object localization, segmentation, and image generation. However, creating an effective prompt requires domain expertise and significant effort. 
Additionally, in practical manufacturing environments, it is typical for data to be unlabeled, which creates difficulties for the prompt-tuning of pre-trained models such as SAM. Therefore, we focus on using unsupervised techniques for creating visual prompts to assist the defect segmentation of the unlabeled XCT images using the SAM inference model.  
  
 In our study, we investigate the utilization of point prompts in SAM and use the location coordinates of the pores that make up the foreground. This is done as follows. As described before in the data preprocessing step, we first select a set of images from a 3D printed sample, run unsupervised clustering i.e., K-means, and K-medoids; to group them in respective clusters, and obtain the centroid or the medoid images. For ease of reference, we call both the centroids and medoids `Centroids' in the rest of the paper given they serve the same purpose in our methodology. We then employ Binary Thresholding (Algorithm \ref{alg1}) to partition each centroid image into the foreground representing defects, and background encompassing the remaining data. Following this we randomly select a small subset of 2D location coordinates from the foreground pixels of these stored centroid images. These selected coordinates are labeled as class `1' to comply with SAM's prompt encoder requirement for foreground labeling. Finally, we apply them as prompt inputs to SAM during the inference. Generating prompts using centroids of clustered images avoids the overhead of creating prompts from selected regions of each individual image on which segmentation is being performed thereby improving the latency for performing real-time segmentation with SAM. Our primary goal is to maintain an unsupervised methodology throughout, both in generating prompts from the data and in executing real-time defect segmentation for images. Below is the pseudocode for our Prompt Generation algorithm.  
\begin{algorithm}[H]
\SetAlgoLined
\textbf{Input:} Centroid Image \(C_k\); \\
\textbf{Output}: 2D point coordinates \(p\), Labels \(l\); \\

\textbf{Steps}: \\
Select the respective centroid images \(C_k\) from cluster \(k\); \\
Apply the Binary Thresholding (Algorithm \ref{alg1}) to \(C_k\); \\
Collect \(p\) = \((x,y)\); \\
where, \((x,y) =  \left\{
\begin{aligned}
& (x \subset (X,Y) \mid X \in \text{X axis coordinates of foreground in \(C_k\)}) \\
& (y \subset (X,Y) \mid Y \in \text{Y axis coordinates of foreground in \(C_k\)})
\end{aligned}
\right\}\)

Create labels \(l = \{1\}^n\) for \((x,y)\), where \(n = |x| = |y|\); \\
\textbf{end}
\caption{Prompt Generation}\label{alg2}
\end{algorithm}

Below is the pseudocode for our porosity segmentation framework using SAM with centroid-based prompt generation by unsupervised clustering. 

\begin{algorithm}[H]
\SetAlgoLined
\textbf{Input:} Clustered XCT images $I_k^P$, Point Prompts $p$;\\
\textbf{Output:} Binary Masks $M_i$, Predicted IoU score $S_i$;\\

\textbf{Steps}: \\
Select the respective centroid images $C_k$ for cluster $k$;\\
Generate point prompts $p$ by Prompt generation Algorithm 2;\\
Input the images $I_k^P$ from cluster $k$ and Point Prompts $p$ to SAM;\\
Set $Thresh = 0.90$; \\
Run SAM inference and obtain $M_i, S_i$ where $0 \leq i \leq 2$;\\
\For{each $M_i, S_i$}{
    Select $M_1, S_1$ (part)\;
    \If{$S_1 > \text{Thresh}$}{
        Select $M_0$, $S_0$ (subpart)\;
    }
}
\caption{Unsupervised Porosity Segmentation Framework}\label{alg3}
\end{algorithm} 

\subsection{Prompt bootstrapping}\label{subsec4.3}

As described in Algorithm \ref{alg2}, prompts are generated for a given image by randomly selecting a subset of points from the corresponding centroid image. To determine the variability of this process and its impact on the final segmentation results we perform bootstrapping of the prompts i.e. the 2D
location coordinates of the foreground of each centroid. Bootstrapping \cite{efron1994introduction} is a technique that has been used for various purposes including estimation of the sampling distribution of a statistic like the median, performing hypothesis testing \cite{efron1992bootstrap}, and generating prediction intervals for future observations \cite{das2021predictive}. The original bootstrap algorithm resamples the entire data with replacement. However, in our case, this is not required as we aim to verify the stability of the segmentation process using a subset of the centroid points corresponding to Class `1'. We apply $m$ out of $n$ bootstrapping \cite{lee2006m} i.e. sampling these points with replacements where $n$ is the total number of centroid points for Class `1' and $m << n$ is the number of resampled points. Full details of the process are described below.  

In our study, for each image, we conduct $m$ out of $n$ bootstrapping on the 2D location coordinates of the foreground of each corresponding centroid where $n$ varies from sample to sample ranging from 1400 to 3,00,000. Thus, $m$ is set to 10,000 for all samples except for Sample 4 due to the significantly lower number of instances, we set $m$ equal to 1000 for this specific case.  For each subsequent iteration of bootstrapping, we randomly select a new set of prompts, with replacement. This means that points selected during the previous resampling step are not removed from the source and can be selected again in the next run. For each of the $m$ resampled sets of prompts, we execute our framework with the SAM inference model and calculate the DSC scores of the predicted masks. We do this for $B= 100$ bootstrap iterations. This generates a distribution of segmentation scores (DSC - refer to Section \ref{subsec5.1}) for every image. Following this we calculate the 95\% confidence intervals over all the images used for segmentation from Samples 3, 4, 5, and 6. These intervals are then used to estimate the variability of our results. Bootstrapping results are discussed in Section \ref{subsec5.2}.

\begin{algorithm}[H]
\SetAlgoLined
\textbf{Input:} Clustered XCT images $I_k^P$, Centroid image \(C_k\); \\
\textbf{Output:} DSC scores, \(D_k^i\); \\

\textbf{Steps}: \\
First, choose the size of prompts, \(m\), where \(m \ll n\) where \(n = |(X,Y)|\); \\
\For{each cluster \(k\)}
{
    Collect \((X,Y)\) from \(C_k\) [Algorithm \ref{alg2}]; \\
    \For{sample $i = 1$ to $B$}
    {
        Randomly sample $p_i$, ($|p_i| = m$); \\
        Run SAM inference model on \(I_k\) and $p_i$; \\
        Calculate DSC scores for predicted masks, \(D_k^i\); \\
        Store DSC scores, \(D_k^i\);
    }
    Compute means and confidence interval of DSC scores, \(D_k^i\);
}
\caption{Bootstrapping Prompts}\label{alg4}
\end{algorithm}

\section{Results}\label{sec5}

We apply the proposed methodology and measure segmentation performance using two different strategies. First, the accuracy of segmentation is measured following the steps as outlined in Algorithm \ref{alg3} using data from Samples 3, 4, 5, and 6. Following this, the variability of the generated output in the previous case is measured following the steps as outlined in Algorithm \ref{alg4}. The details are as described below.

\subsection{Performance Analysis using Centroid-Based Prompts}\label{subsec5.1}

We select a set of XCT images from each sample and partition them into three clusters using unsupervised clustering methods, namely K-means and K-medoids. Next, we implement our framework (Algorithm \ref{alg3}) on the four samples separately. To ensure fairness in comparison, we maintain equal-sized selected sets for all samples. While employing K-medoid clustering on the XCT images, we utilize both Euclidean distance and Dynamic Time Warping (DTW) distance as distinct metrics for effective comparison of the resulting medoids. Notably, the medoid images obtained from K-medoid clustering are identical when using either distance metric. 

We use the individual Reference Binary mask obtained for each image as outlined in Algorithm \ref{alg1} to evaluate the results from the centroid-based prompt generation technique. We calculate the Dice Similarity Coefficient (DSC) \cite{carass2020evaluating} between the Reference Binary masks and the predicted masks by our framework to measure the accuracy of segmentation. DSC measures the similarity between two binary masks by comparing their spatial overlap (equation \ref{eq11}). DSC score ranges from 0 to 1, where DSC = 1 meaning the perfect overlap, and DSC = 0 meaning no overlap at all between the masks. 
\begin{equation}
\label{eq11}
    DSC(M, R) = \frac{2 \times |M \cap R|}{|M| + |R|}
\end{equation}
Here, \(M\) = Predicted mask by our framework, \(R\)= Reference Binary mask obtained by Thresholding, \( |M| \) and \( |R| \) are the sizes (cardinalities) of sets \(M\) and \( R \), and \( |M \cap R| \) is the number of elements common to both sets \( M \) and \( R \) respectively. 
We demonstrate the comparison among the DSC scores obtained from the prediction by SAM with different prompt generation techniques in Table \ref{tab2}. 

\begin{table}[h]
\captionsetup{width=0.8\textwidth}
\centering
\small
\begin{tabularx}{\textwidth}{@{}c *{5}{>{\centering\arraybackslash}X}@{}}
\hline
\multirow{2}{*}{\textbf{Samples}} & \multicolumn{4}{c}{\textbf{Dice Similarity Score}} \\ \cline{2-5} 
                           & \textbf{K-means} & \textbf{K-medoids} & \textbf{Reference Masks} & \textbf{No Prompts} \\
\hline
\textbf{Sample 3}            & $0.62\pm0.034$         & $0.62\pm 0.036$        & $0.63\pm 0.029$  & $0.0019\pm 0.0006$ \\
\textbf{Sample 4}            & $0.436\pm0.12$      & $0.35\pm 0.13$        & $0.37\pm 0.11$   & $0.0299\pm 0.0081$ \\
\textbf{Sample 5}            & $0.88\pm0.019$     & $0.88\pm0.015$      & $0.88\pm 0.016$    & $0.0022\pm 0.0021$  \\
\textbf{Sample 6}            & $0.70\pm0.09$     & $0.75\pm0.04$   & $0.70\pm 0.07$   & $0.0002\pm 0.0002$  \\

\hline
\end{tabularx}
\caption{Mean Dice Similarity Score obtained from the different prompt generation techniques on all the samples. The standard deviations are calculated to show the spread of the Dice Similarity Scores, indicating the variability of the results obtained from different prompt generation techniques across all samples. The first two columns show the DSC scores achieved using centroids derived from different clustering techniques respectively. The next column represents the DSC scores obtained from prompts generated by using the respective Reference Binary masks. The final column displays the DSC scores obtained without using any prompts. }
\label{tab2}
\end{table}
Without any prompts, SAM returns ambiguous masks of the defects with very low DSC values shown in Table \ref{tab2}. Moreover, we observe that the highest DSC scores are achieved with prompt sizes equal to or higher than 10,000 of the total pixels of defect regions. In addition, it should be noted that when we apply our framework directly to original XCT images,  the obtained results are often ambiguous, and the optimal prompt size required for improved segmentation varies across different samples.  

\begin{figure}[H]%
\centering
\includegraphics[width=1\textwidth]{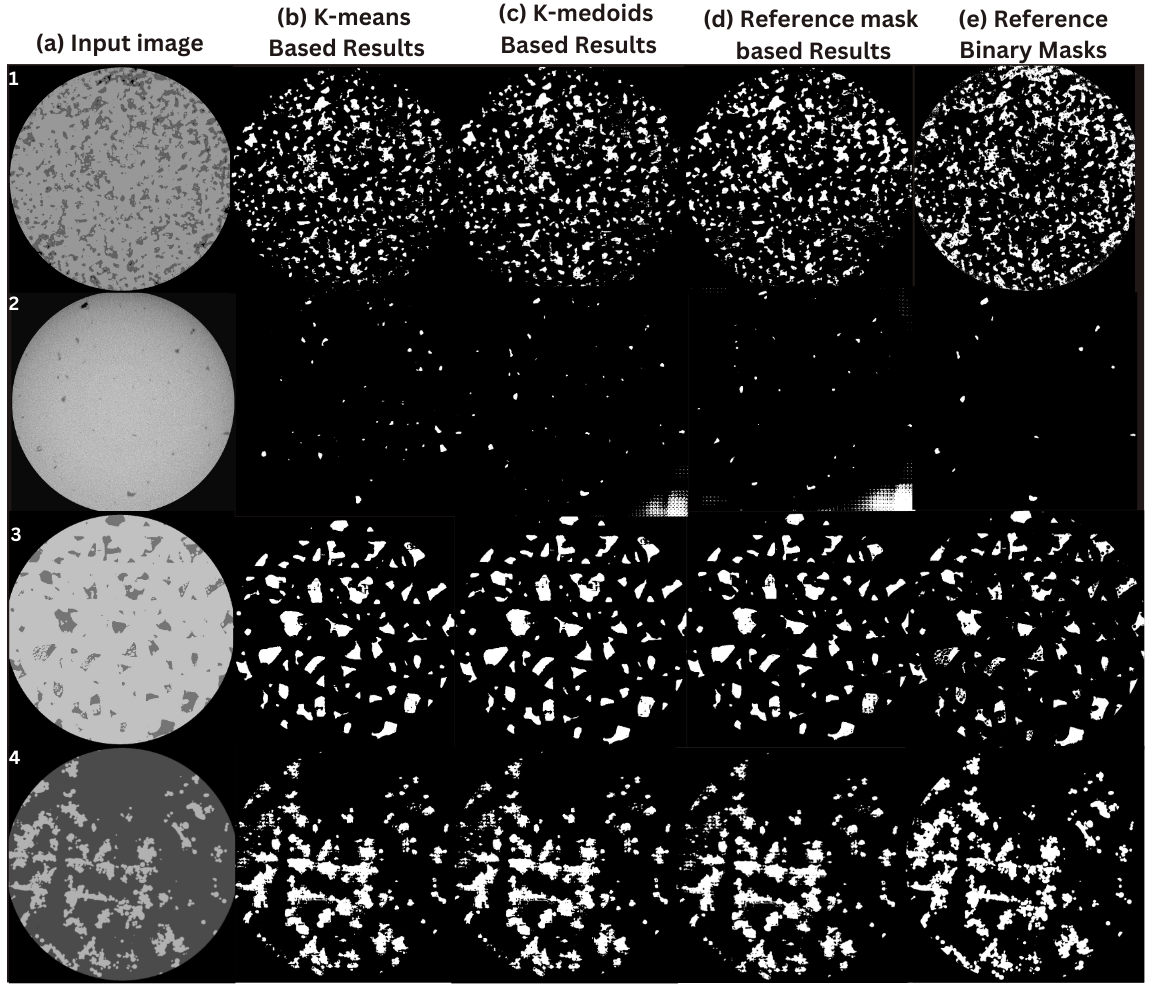}
\caption{Prediction results by our framework. Rows 1,2,3,4 represent examples of XCT images from all the samples. (a.1) Input XCT Image from Sample 3, (b.1) Predicted mask by our framework using K-means-based prompts, (c.1) Predicted mask by our framework using K-medoids-based prompts, (d.1) Predicted mask by our framework using corresponding Reference Binary Masks-based prompts,(e.1) Corresponding Reference Binary mask, (a.2) Input XCT Image from Sample 4, (b.2) Predicted mask by our framework using K-means-based prompts, (c.2) Predicted mask by our framework using K-medoids-based prompts, (d.2) Predicted mask by our framework using corresponding Reference Binary Masks-based prompts,(e.2) Corresponding Reference Binary mask, (a.3) Input XCT Image from Sample 5, (b.3) Predicted mask by our framework using K-means-based prompts, (c.3) Predicted mask by our framework using K-medoids-based prompts, (d.3) Predicted mask by our framework using corresponding Reference Binary Masks-based prompts,(e.3) Corresponding Reference Binary mask, (a.4) Input XCT Image from Sample 6, (b.4) Predicted mask by our framework using K-means-based prompts, (c.4) Predicted mask by our framework using K-medoids-based prompts, (d.4) Predicted mask by our framework using corresponding Reference Binary Masks-based prompts,(e.4) Corresponding Reference Binary mask.}\label{fig4.1}
\end{figure}

For Samples 3, 5, and 6, we utilize a prompt size of 10,000 points of location coordinates of total pixels belonging to the defect regions. However, in the case of Sample 4, where the number of pixels corresponding to defects is significantly lower, we opt for a sample size of approximately 1,000 points of location coordinates. Our framework demonstrates effective performance in segmenting pores within Sample 5, characterized by moderate pore dimensions and uniform pore distribution. In the case of Sample 6, featuring irregularly shaped and large-to-medium-sized pores, and Sample 3, exhibiting a moderately intricate distribution of smaller-sized pores, the framework provides slightly reduced yet acceptable results. However, its performance is less satisfactory in the case of Sample 4. Here, both the size and number of pores are very small. Additionally, a subset of pixel values within non-defective regions closely resembles those associated with porous regions. This arises from the XCT imaging of densely solid, defect-free regions further complicating segmentation challenges. We observe that prediction accuracies yielded by both clustering methods, K-means, and K-medoids, as well as the Reference Binary mask-based prompts, fall within a similar range across all samples.  Since the prompts generated from the Reference Binary masks have the exact locations for the individual images compared to the centroids or medoids this suggests that the unsupervised clustering-based method of prompt generation provides a more generalized and efficient approach.  Prompt generation using the Reference Masks is impractical in real-life scenarios, as it is highly unlikely for LAM data to possess annotated labels. Moreover, this method requires using separate Reference Binary masks for every individual image which hinders performing segmentation in real-time. In contrast, centroid-based prompts offer faster generation with comparable performance as shown in the results. For reference, the best predictions obtained for each of the samples using our framework are demonstrated in Figure \ref{fig4.1}. 

To verify the robustness of our approach we use the centroids obtained from images of previous layers to perform segmentation on new images from successive ones. 
We use Sample 5 for this investigation owing to its superior performance as discussed above. In this case, we observe a slight decrease in the DSC score $0.83\pm0.09$ compared to the accuracy obtained on previous layers. This is expected since the centroid images from the previous layer's data do not possess the exact information on the location of the pores for the next layers. However, it can be observed that by maintaining the same set of process parameters during the sample printing, the average morphology and distribution of the pores over the height for a certain sample do not change significantly except for their locations in the layers. This approach of using centroids of previous layers to generate prompts for successive ones can be useful in case of trading off between the accuracy of defect segmentation and the latency associated with regenerating a set of prompts using data from the new layers.

\subsection{Performance Analysis using Bootstrapping}
\label{subsec5.2}
To estimate the variability of our segmentation results using a random set of coordinates from the centroids as prompts, we utilize Algorithm 4, which involves the application of bootstrapping the prompts for all four samples. To do this, we use centroids from K-means clustering. The results from Kmedoids clustering are expected to be on similar lines. 

\begin{figure}[H]%
\centering
\includegraphics[width=0.8\textwidth]{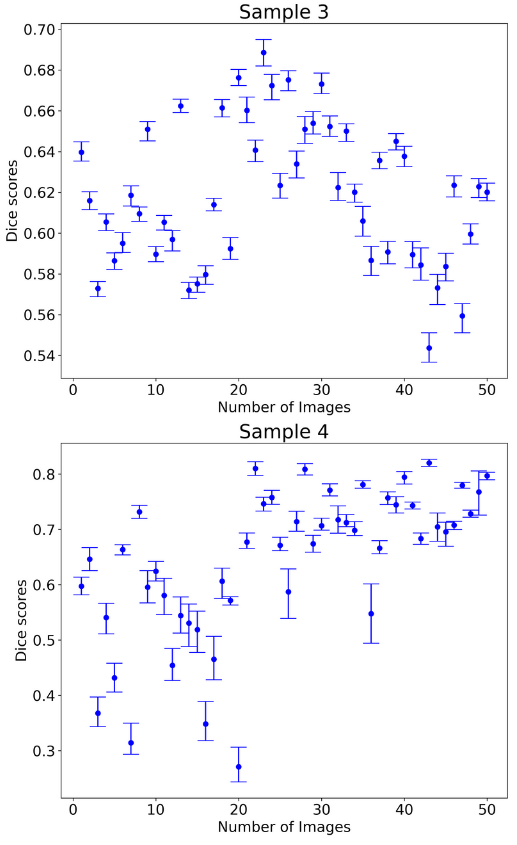}
\caption{ The 95\% confidence interval based on the quantile of the distribution of DSC scores for the predicted masks generated by our framework for Sample 3 and Sample 4 obtained from Bootstrapping of the prompts.}\label{fig5.3}
\end{figure}

\begin{figure}[H]%
\centering
\includegraphics[width=0.8\textwidth]{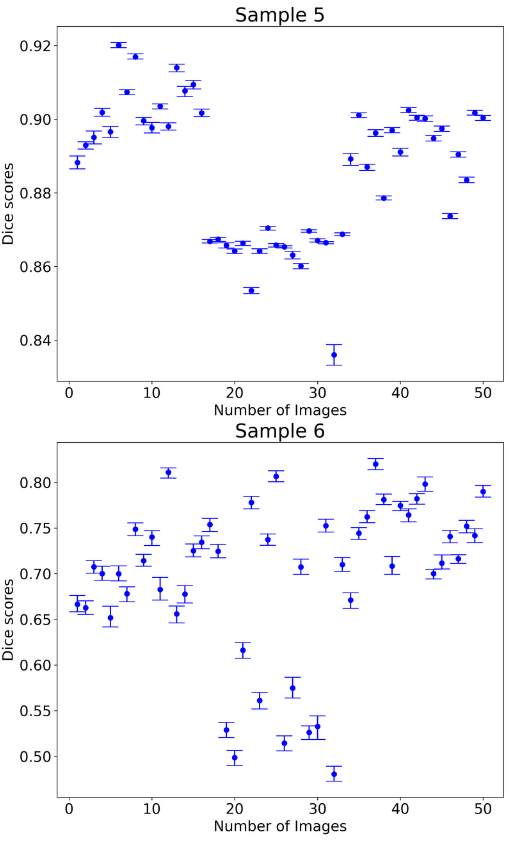}
\caption{ The 95\% confidence interval based on the quantile of the distribution of DSC scores for the predicted masks generated by our framework for Sample 5 and Sample 6 obtained from Bootstrapping of the prompts.}\label{fig5.4}
\end{figure}

Using the distributions estimated through bootstrapping the prompts, we generate 95\% confidence intervals based on the quantiles \cite{hoeffding1994probability} of the estimated distribution. We then calculate the mean lengths of these intervals across all four sets of samples. The results are shown in Table \ref{tab4}. The 95\% confidence intervals for all the samples are shown in Figures \ref{fig5.3} and \ref{fig5.4} respectively. We observe that except for Sample 4, the intervals are relatively tight thereby indicating a low variability by using centroid-based randomly sampled prompts.

\begin{table}[h]
\captionsetup{width=0.8\textwidth}
\centering
\small
\begin{tabularx}{\textwidth}{@{}l*{1}{>{\centering\arraybackslash}X}@{}}
\hline
\textbf{Samples} & \textbf{Mean length} \\
\hline
\textbf{Sample 3} & $0.0103 \pm 0.0025$ \\
\textbf{Sample 4} & $0.0396 \pm 0.0236$ \\
\textbf{Sample 5} & $0.0017 \pm 0.0009$ \\
\textbf{Sample 6} & $0.0153 \pm 0.0034$ \\
\hline
\end{tabularx}
\caption{Mean length and standard deviation of the 95\% confidence intervals of the distributions estimated by Bootstrapping using quantiles of the estimated DSC distribution across all the samples.}
\label{tab4}
\end{table}

\section{Discussion}\label{sec6}
\begin{figure}[!ht]%
\centering
\includegraphics[height=4.25 in,keepaspectratio]{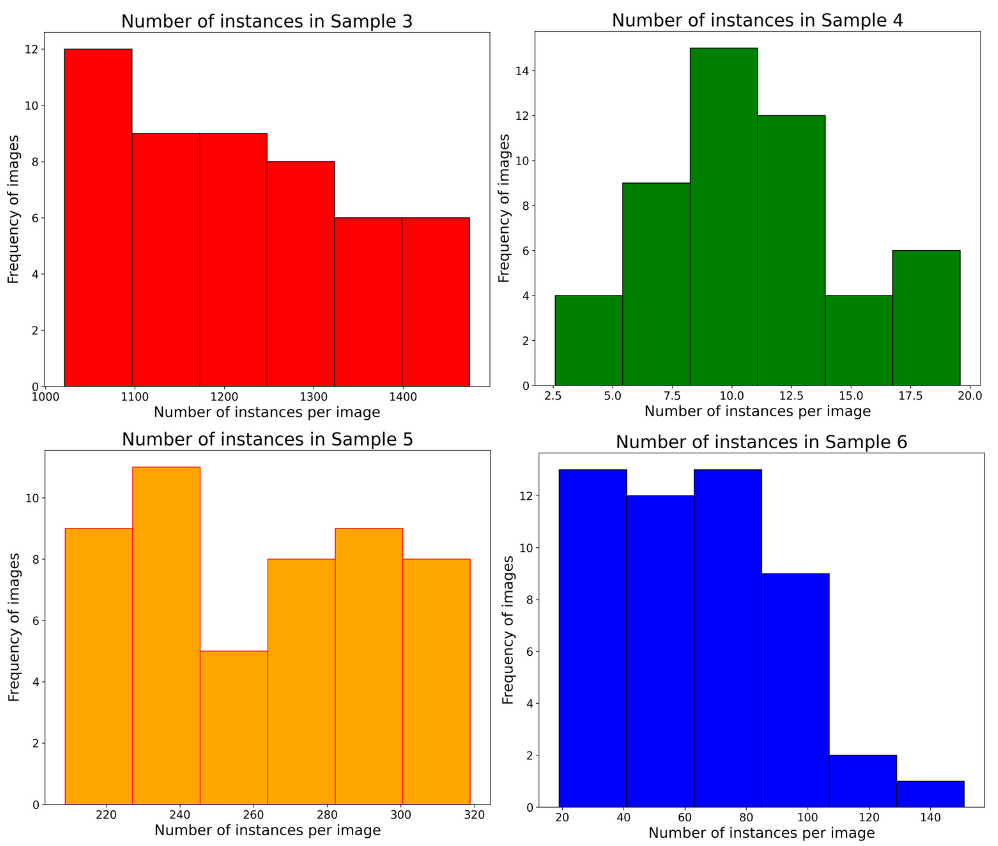}

\caption{Number of instances per XCT image in Samples 3, 4, 5, and 6 respectively.}\label{fig5.1}
\end{figure}

The quality and accuracy of semantic segmentation results can be influenced by the number of instances, their size, and their distribution within an image. This complexity is especially distinct when multiple instances of similar objects are present, making the differentiation very challenging. \cite{amirkhani2024visual} showed that it may be convenient to perform segmentation on a dataset with a limited number of instances, while a more diverse dataset with a broader distribution suggests a moderate level of complexity for segmentation. Furthermore, datasets with heavy-tailed distributions present additional challenges, reflecting greater diversity and complexity in segmentation tasks. We visualize the approximate distribution of instance (pore) counts per image in our case study datasets based on connected components \cite{rosenfeld1966sequential} in Figure \ref{fig5.1}. 

We observe that our framework achieves promising results for samples with low to moderate instance counts and relatively uniform distributions, exemplified by sample 5 which exhibits a high DSC score of \((0.88\pm0.09)\). On the other hand, this framework encounters challenges when dealing with datasets with heavy-tailed distributions such as Sample 6 and higher instance counts such as Sample 3. In the case of Sample 3, in addition to the high pore count with a highly complex spatial distribution, we also encounter the presence of minute trapped powder particles within a few pores, which are marked as the non-defect background in the Reference Binary masks during thresholding. In this case, misclassifications i.e. segmentation errors occur because their pixel intensities closely resemble those of the non-defect solid surface (Figure \ref{fig4.1}).
In the case of Sample 6, our framework exhibits limited success by segmenting the non-defect regions instead of the pores. This outcome may be attributed to the unique pore morphology observed in this sample, characterized by large size, irregular shapes, and interconnectedness. It is important to highlight that, for this particular sample, we have generated the Reference masks based on non-defective regions rather than exclusively focusing on the defect areas.  This methodology has been adopted to ensure a comprehensive evaluation of the accuracy of our framework. 

Another limitation of our framework is observed in the context of Sample 4, where the segmentation outcomes exhibit significant fluctuations. This is evidenced by the higher standard deviation, indicating greater variability, as shown in Table \ref{tab2}. Although the instance count is low for this particular sample, the variability can be attributed to the presence of variations and noise within the background pixels at a granular level specific to this sample. Additionally, a number of pixels belonging to the non-defect area have a similar pixel intensity as the pores, contributing to the observed fluctuations. Furthermore, we note that, for Sample 4, the most accurate results are obtained from the masks segmenting the sub-parts (Algorithm \ref{alg3}) from the multi-mask outputs of SAM. This can be attributed to the limited number of pores with very small sizes to segment in this sample. SAM fails to recognize these pores in both part and whole object returning masks, resulting in parts and objects being returned indistinguishable. 

In summary, our framework has demonstrated excellent segmentation performance with Sample 5, and moderate performance with Samples 3 and 6 respectively while encountering challenges with Sample 4. Previous studies \cite{ouidadi2023defect} have also reported similar challenges, noting that pre-trained U-Net models struggle with background noise in Sample 4.

\section{Our Vision}\label{sec7}
\begin{figure}[H]%
\centering
\includegraphics[keepaspectratio, width=1\textwidth]{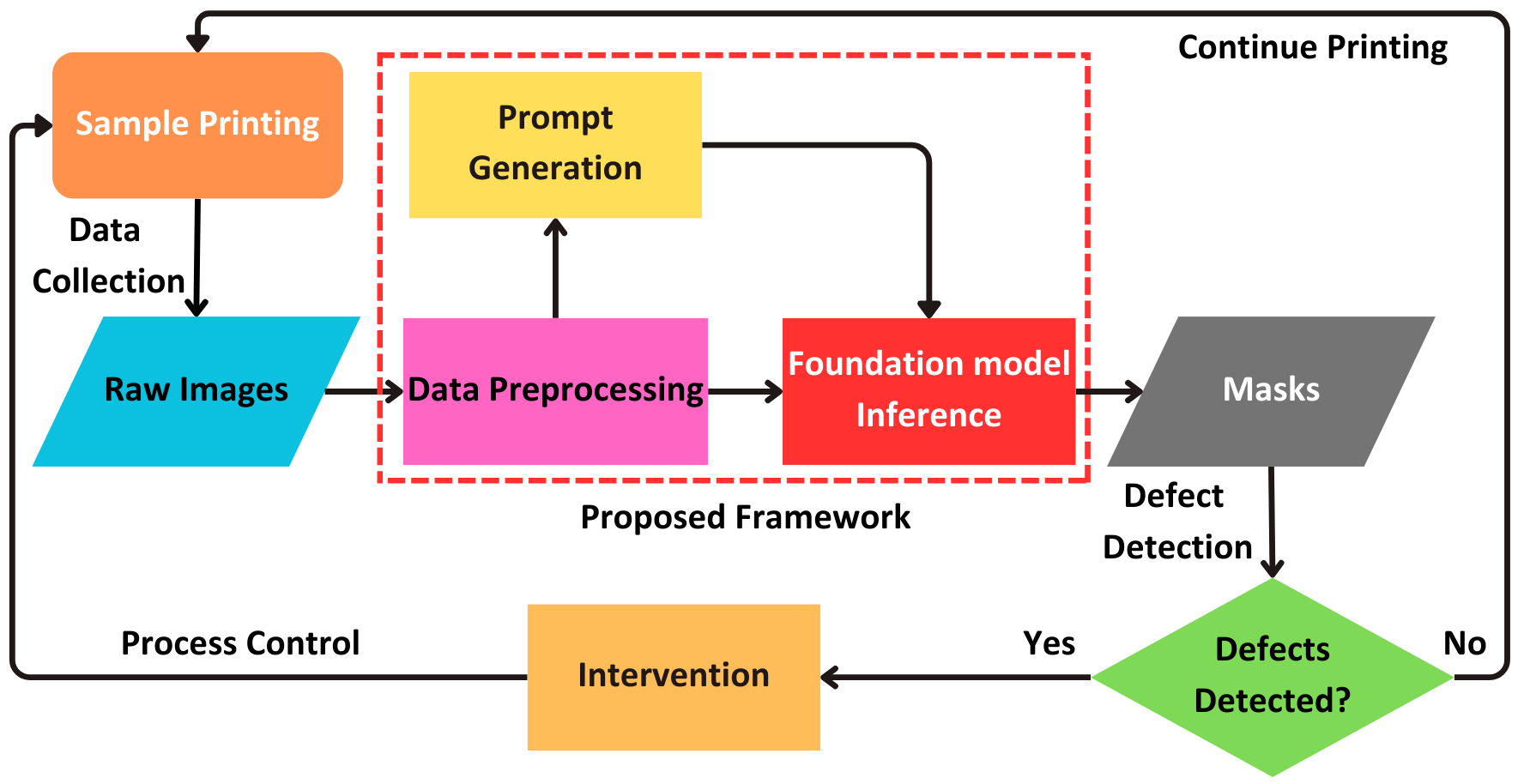}

\caption{Anomaly Detection Pipeline with Foundation Model Inference}\label{fig6.1}

\end{figure}

Defect detection has been typically performed by either applying user-determined thresholds limits \cite{guo2022unsupervised, yi2020patch} to determine defect counts, defect sizes or volumes on account of specific mechanical properties like tensile strength required for end products \cite{thompson2016x, zhang2022nondestructive}, or by using Machine Learning classifiers \cite{chauhan2014comparison, alqahtani2020machine, gobert2020porosity, bellens2022evaluating}. As mentioned earlier, integrating AI frameworks into LAM involves serious challenges such as process latency and the absence of labels, necessitating extensive human intervention. Toward this end, our centroid-based prompt generation methodology for performing defect segmentation with the zero-shot generalization capability \cite{kirillov2023segment} of SAM enables a simple and seamless framework that can minimize human involvement. The next step is to implement this framework in the LAM system for real-time application. Leveraging foundation models like SAM for real-time defect detection shows potential for real advancement over existing methodologies in LAM domains. For a single RGB image of size \(1010\times984\) pixels using multiple point prompts as described in Algorithm 4, the total run time to generate three valid masks ranges from 3.8 to 4.4 seconds. This runtime is well-suited to the rapid layer-wise printing characteristic of LAM processes. Thus, we envision a real-time defect detection pipeline in LAM incorporating our framework to establish an autonomous process. Given LBAM's layer-by-layer printing strategy, we can collect the data of the first few layers and then run our framework to store the centroids for prompt generation and perform defect detection. Then, we continue to collect real-time data and use the stored centroids to detect any presence of defects using our framework as shown in Figure \ref{fig6.1}. With the advent of powerful edge computing platforms \cite{shi2016edge} using Graphics Processing Units (GPUs) or Field Programmable Gate Arrays (FPGAs) \cite{colbert2021competitive} that can be seamlessly integrated with the Internet of Things (IoT) sensors \cite{kumar2019internet} in the system, it is possible to enable faster inference in order to enhance operational efficiency along with uninterrupted communication between devices e.g., cameras, deployment interfaces, and production systems. 

However, there exists a challenge of process latency during data collection and real-time process control in printing \cite{mani2017review}, which current technologies are struggling to overcome. Thus, finding the right balance between the volume of data, the complexity of integrated AI frameworks with inference models, and the need for real-time deployment becomes the next crucial step in additive manufacturing research.  

\section{Conclusions}\label{sec8}

In the realm of LAM processes, XCT imaging is essential for detecting inter-layer defects in printed samples. Image segmentation, one of the most critical steps in this process, faces significant challenges such as the scarcity of labeled XCT images, extensive fine-tuning requirements, and substantial human intervention. To address these challenges, in this work, we introduce a novel porosity detection framework that leverages the robust capabilities of the Segment-Anything foundation model, which has recently revolutionized image segmentation research. Our framework employs an unsupervised approach to generate contextual prompts directly from XCT data. The proposed method has demonstrated a high accuracy rate of higher than 80\% in segmenting defects within XCT images of L-PBF-printed part layers, validating its effectiveness. We provide a comprehensive explanation of our methodology and discuss our findings and inherent limitations. Additionally, we share valuable insights from experiments conducted across various L-PBF samples, showcasing diverse pore morphologies and distributions. We consider our study a foundational effort that advances future research by extending the potential of promptable foundation models such as SAM in the manufacturing domain. Nevertheless, the use of a relatively large number of multi-point prompts in our specific case, may require optimization within the embedding space \cite{huang2024learning} through few-shot training. These findings highlight the necessity of enhancing our unsupervised prompt generation approach with a prompt optimization strategy, either in the original or embedded space, to improve the efficiency of prompt utilization in our framework. To address this, we plan to explore latent or embedded spaces and integrate an efficient prompt optimization strategy using a few-shot prompt learning technique. This enhancement aims to improve the overall unsupervised prompt generation framework for LAM data in the future.

\section*{Declarations}

\textbf{Funding:} This material is based upon work supported by the National Science Foundation under Grant No. 2119654. Any opinions, findings, conclusions, or recommendations expressed in this material are those of the author(s) and do not necessarily reflect the views of the National Science Foundation. 

\textbf{Conflict of interest/Competing interests:} As authors of this work, we declare that we have no conflicts of interest.

\textbf{Availability of data and materials:} The dataset used and analyzed during the current study are available in NIST Engineering Laboratory's repository, \url{https://www.nist.gov/el/intelligent-systems-division-73500/cocr-am-xct-data}

\textbf{Code availability:} The underlying code for this study is not publicly available but can be made available on reasonable request to the corresponding authors.  

\textbf{Authors' contributions: }IZE: Data Preprocessing; Code and Methodology, Formal Analysis, Validation, Original Draft; IA: Conceptualization, Methodology, Discussion, Supervision; ZL: Conceptualization, Supervision, Domain Knowledge, Research Funding; SD: Conceptualization, Methodology and Statistical Analysis, Discussion, Supervision, Computational Resources. All authors contributed to manuscript editing and revision. 

\textbf{ Acknowledgments:} The authors would like to acknowledge the Pacific Research Platform, NSF Project ACI-1541349, and Larry Smarr (PI, Calit2 at UCSD) for providing the computing infrastructure used in this project.

\bibliographystyle{elsarticle-num-names}
\bibliography{reference}

\end{document}